\documentclass[acmtog]{acmart}
\acmSubmissionID{572}

\usepackage{booktabs}
\usepackage{cleveref}  
\usepackage{multirow}
\usepackage{array}

\usepackage{graphicx}

\usepackage[ruled]{algorithm2e} 

\SetAlFnt{\small}
\SetAlCapFnt{\small}
\SetAlCapNameFnt{\small}
\SetAlCapHSkip{0pt}

\acmJournal{TOG}

\newcommand{\method}{LayerFlow\xspace}

\copyrightyear{2025}
\acmYear{2025}
\setcopyright{acmlicensed}\acmConference[SIGGRAPH Conference Papers '25]{Special Interest Group on Computer Graphics and Interactive Techniques Conference Conference Papers }{August 10--14, 2025}{Vancouver, BC, Canada}
\acmBooktitle{Special Interest Group on Computer Graphics and Interactive Techniques Conference Conference Papers (SIGGRAPH Conference Papers '25), August 10--14, 2025, Vancouver, BC, Canada}
\acmDOI{10.1145/3721238.3730662}
\acmISBN{979-8-4007-1540-2/2025/08}

\begin{document}
\title{\method: A Unified Model for Layer-aware Video Generation}

\author{Sihui Ji}
\authornote{Work is done when Sihui Ji worked as interns in Alibaba DAMO Academy. 
}
\orcid{0009-0004-8552-4985}
\affiliation{%
 \institution{The University of Hong Kong, DAMO Academy, Alibaba Group}
 \country{China}
 }
\email{sihuiji.cs@connect.hku.hk}
\author{Hao Luo}
\affiliation{%
 \institution{DAMO Academy, Alibaba Group, Hupan Laboratory}
 \city{Hang Zhou}
 \postcode{310023}
 \country{China}
}
\email{michuan.lh@alibaba-inc.com}
\author{Xi Chen}
\affiliation{%
\institution{The University of Hong Kong}
\country{Hong Kong}
}
\email{xichen.csai@connect.hku.hk}
\author{Yuanpeng Tu}
\affiliation{%
 \institution{The University of Hong Kong}
 \country{Hong Kong}
}
\email{yuanpengtu@connect.hku.hk}
\author{Yiyang Wang}
\affiliation{%
 \institution{The University of Hong Kong}
 \country{Hong Kong}
 }
\email{yiyangwang@connect.hku.hk}
\author{Hengshuang Zhao}
\authornote{Corresponding author.}
\affiliation{%
 \institution{The University of Hong Kong}
 \country{Hong Kong}
}
\email{hszhao@cs.hku.hk}

\begin{abstract}
We present \method, a unified solution for layer-aware video generation. Given per-layer prompts, \method generates videos for the transparent foreground, clean background, and blended scene. It also supports versatile variants like decomposing a blended video or generating the background for the given foreground and vice versa. 
Starting from a text-to-video diffusion transformer, we organize the videos of different layers as sub-clips, and leverage layer embeddings to distinguish each clip and the corresponding layer-wise prompts. 
In this way, we seamlessly support the aforementioned variants in one unified framework.  
For the lack of high-quality layer-wise training videos, we design a multi-stage training strategy to accommodate static images with high-quality layer annotations.   
Specifically, we first train the model with low-quality video data.
Then, we tune a motion LoRA to make the model compatible with static frames. 
Afterward, we train the content LoRA on the mixture of image data with high-quality layered images along with copy-pasted video data. 
During inference, we remove the motion LoRA thus generating smooth videos with desired layers. 
\end{abstract}

%
%




%
%



\begin{teaserfigure}
\includegraphics[width=\textwidth]{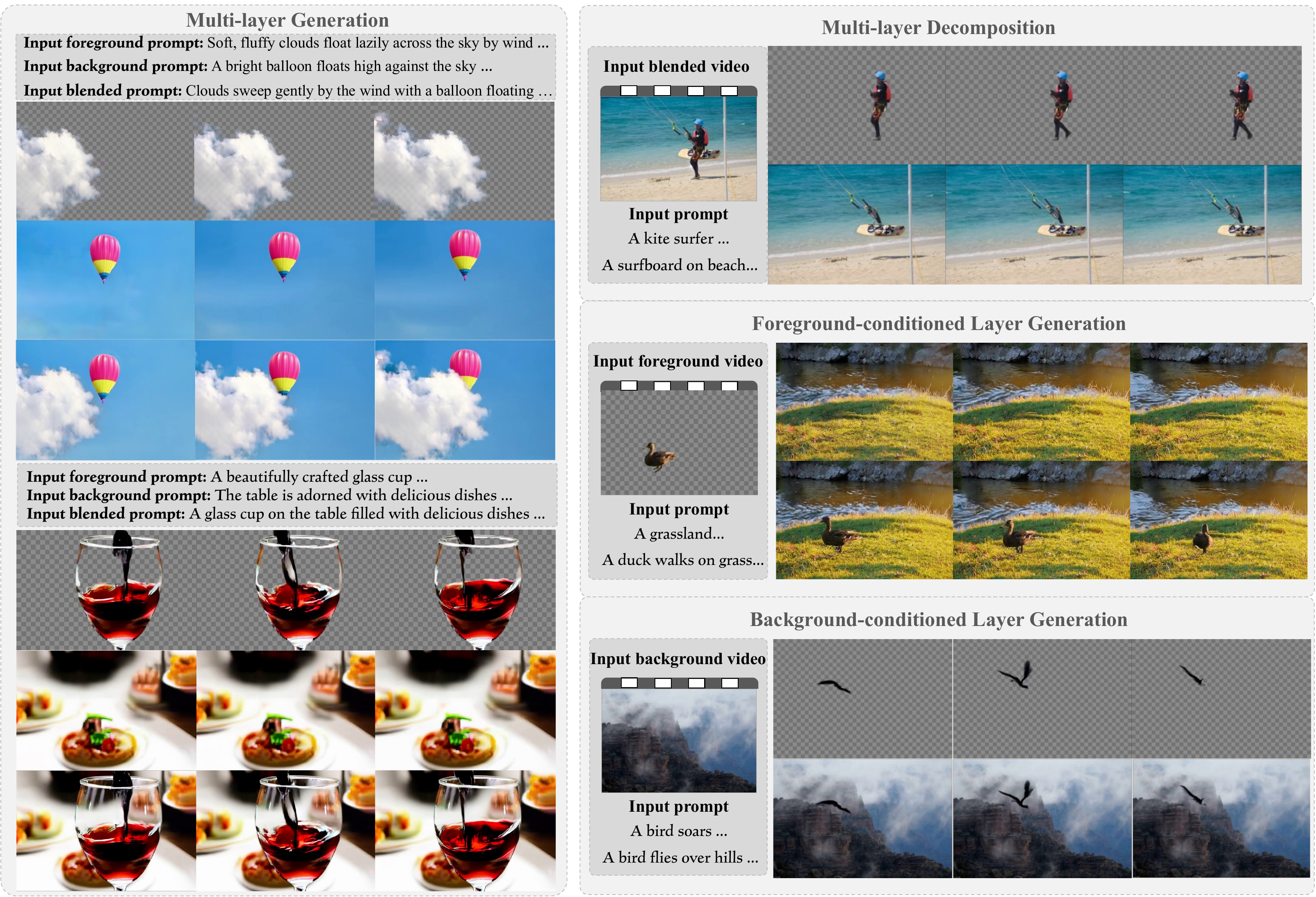}
\caption{\textbf{Demonstration for the applications of \method.} Given layer-wise prompts, our method produces videos for a transparent foreground, a clean background, and a blended scenario. It also supports different user-provided conditions, enabling users to decompose and recompose videos creatively.
}
\label{fig:teaser}
\end{teaserfigure}

\maketitle

\section{Introduction}
\label{sec:intro}
Video generation is a burgeoning topic with broad real-world applications~\cite{hong2022cogvideo,brooks2024video,guo2023animatediff}. Previous works tackle text-to-video (T2V) generation with various approaches, 
and make exciting advancements~\cite{yang2024cogvideox,chen2024videocrafter2}.  Some works further explore enhancing the controllability of the generated videos with sequential inputs like optical flows, or depth maps, bringing more convenience and imagination to video content creation~\cite{yang2024direct,ma2023trailblazer,wang2024motionctrl}. 
In this work,  we focus on layer-aware video generation, which means the simultaneous synthesis of foreground, background, and blended videos, with the prompt of each layer.

%
Layer-wise generation has potential to support flexible decomposition and recomposition of visual assets, and independent editing in layer level. Furthermore, the inclusion of foreground layer with transparency channel allows realistic effects to be seamlessly integrated into background, benefiting visual production workflows and applications.
In the field of layered image synthesis, several progress has been made like  
LayerDiffuse~\cite{zhang2024transparent} and Alfie~\cite{quattrini2024alfie}, which investigate text-to-RGBA image with transparent effects represented by alpha channel. 
However, layer-wise video synthesis remains relatively underdeveloped attributed to two primary challenges.
%
First, representations of different layers and alpha mattes in videos remain largely unexplored. The inclusion of temporal dimension of videos increases the complexity of incorporating transparency channels.
Second, the scarcity of multi-layer video data significantly hinders progress. 
High-quality layerd video datasets are rare and difficult to construct, posing a barrier to the generalization and diversity of generated videos.
%
As a result of these two difficulties,
there are currently no accessible works synthesizing videos of multiple layers to the best of our knowledge.

Facing the aforementioned challenges, in this work, we propose \method, which supports generating independent layers of a transparent foreground, a clean background and a compositional scenario.
To represent different video layers efficiently, the video segment of each layer including alpha channel, and their corresponding prompts are separately stitched together into a long sequence. Layer embeddings are also inserted into the sequence to endow the model with layer-awareness. Equipped with such a framework, multi-layer videos with transparency can be synthesized simultaneously.
The scarcity and challenges in creation of compliant training dataset force us to devise an effective training pipeline to make full use of accessible data, thus we propose a three-stage training strategy based on two well-designed LoRAs~\cite{hu2021lora} to allow joint image-video data training.
To be specific, first, we finetune the pretrained text-to-video model~\cite{yang2024cogvideox} with video data roughly made by inpainting for initial ability of layer-aware generation. 
Second, Motion LoRA is trained on static multi-layer videos made by copy and paste for adapting the model to image data.
Thus at the final stage, we tune Content LoRA with joint image-video data by turning on or off Motion LoRA according to the type (static or dynamic) of input training data.
During inference, Content LoRA is applied for refining layer-aware generation quality while Motion LoRA is removed for restoring video dynamics.
By borrowing knowledge from high-quality images
and preserving the motion prior through videos, we are capable of 
generating layered videos including transparent foreground and undisturbed backgrounds.

\begin{figure*}[t]
\centering 
\includegraphics[width=1\linewidth]{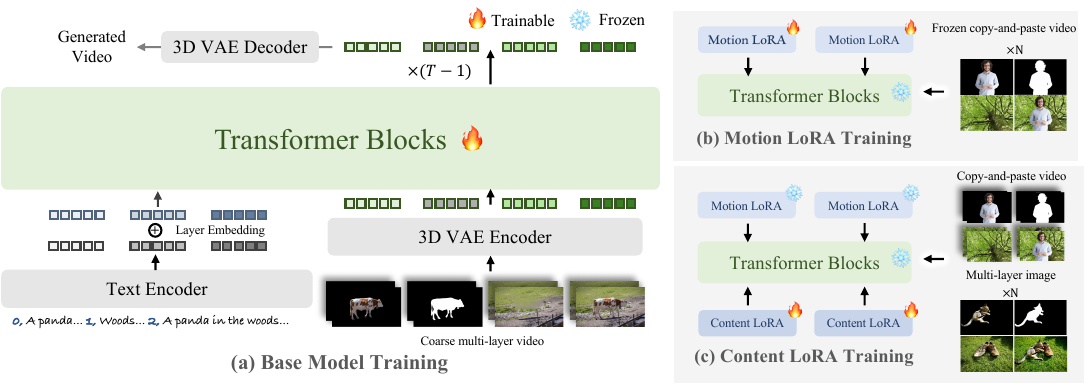} 
\caption{%
   \textbf{Overall pipeline of \method}, which allows for the production of multi-layer videos including transparent foreground, undisturbed background and blended sequences. We organize videos of different layers as sub-clips and concatenate them to form a whole sequence to be encoded by VAE encoder. At the same time, index modification is conducted before prompts are processed by the $T5$ encoder, then layer embedding is added to text embeddings to impart layer awareness.  All the visual patches and text embeddings are fed into transformer blocks as a long tensor. In the process of training, a base model is firstly trained on crudely made multi-layer video data for initial layered generation ability. Motion LoRA tuning prepares the model with accommodations for frozen video and Content LoRA can then borrow knowledge from both high-quality duplicated multi-layer images and copy-pasted video data for improving layer-aware synthesis quality as well as maintenance of motion dynamics.
}
\label{fig:pipeline}
\end{figure*}

\method demonstrates 
promising abilities in both generation quality and semantic alignment over other solutions such as trained solely on videos or generating then animating. Moreover, by removing the noise from the specified video clip, we can condition on that clip for generating the remaining segments, thus achieving multiple derivative applications in~\cref{fig:teaser} in a unified framework. For example, background-conditioned generation draws foreground on the input video 
and vice versa, while multi-layer decomposition subtracts independent video layer from a given video.
\method shows the potential to act as a foundational solution for layered video creation with multi-modal conditions to energize more fancy applications.
\section{Related Work}
\label{sec:related_works}
\paragraph{Video generation and editing.}
The fast-paced growth of T2V models has been phenomenal driven by both the
Transformer architecture~\cite{vaswani2017attention} and diffusion model~\cite{ho2020denoising}. 
Conventional diffusion-based T2V models usually start from pre-trained text-to-image (T2I) models or leverage
large-scale image datasets for training. AnimateDiff~\cite{guo2023animatediff} inflates T2I models by training
a plug-and-play motion module to learn transferable motion priors from real-world videos. 
Using Transformers as the backbone of diffusion models (DiT) ~\cite{peebles2023scalable} has shown great promise since Sora~\cite{brooks2024video} 
presented impressive performances.
CogVideoX~\cite{yang2024cogvideox} is a novel DiT-based model, achieving
long-term consistent video generation with dynamic plots. 
Existing works have also explored video editing. AVID~\cite{zhang2024avid} conducts inpainting on videos of any duration following a similar architecture as AnimateDiff~\cite{guo2023animatediff}, and UniEdit~\cite{bai2024uniedit} proposes a unified tuning-free framework for video motion and appearance editing.
All the above approaches fail to generate layered videos,
while our method
focuses on layer-aware video creation with all related applications.

\paragraph{Layered content generation.} Layered image or video generation is an emerging and challenging topic. LayerDiffuse~\cite{zhang2024transparent} encodes alpha channel transparency into latent space of a pre-trained latent diffusion model for the generation of single transparent images or multiple transparent layers.
Alfie~\cite{quattrini2024alfie} modifies the inference behavior of a pre-trained DiT  
for fully automated obtaining of RGBA illustrations.
TransPixar~\cite{wang2025transpixeler} extends pretrained video diffusion models to generate RGBA videos by jointly modeling RGB and alpha channels through alpha-aware tokens and LoRA-based fine-tuning, while TransAnimate~\cite{chen2025transanimate} combines pre-trained transparent image models with temporal modules and introduces motion-guided control mechanism for controllable RGBA video generation.
Several works~\cite{lee2024generative, gu2023factormatte, lu2021omnimatte} also explored generative layered video decomposition, aiming to separate objects and their associated effects into multiple layers. 
However, there are no available methods of omni solution for various simultaneous multi-layer video generation tasks to the best of our knowledge, and our work
manages to unify solutions of layered video generation and its derived applications in a single framework.

\section{Methods}
\label{sec:methods}

The pipeline of \method is demonstrated in ~\cref{fig:pipeline}. Given three text inputs that seperately describe content of the target layer, our model can generate foreground, alpha channel, background, and blended video with high fidelity and per-layer prompt alignment. 
We first give a brief introduction to the overall framework in \cref{sec:overall_framework}. Following that, our comprehensive training pipeline is outlined in \cref{sec:training} on training strategy and dataset organization. 
\subsection{Overall Framework}
\label{sec:overall_framework}
We first adapt a DiT-based T2V model to a layer-aware generation setting by concatenating sub-clips of all layers as a whole, then align visual contents of each layer with corresponding text prompts by insertion of layer embeddings on layer-wise text embeddings. 
Based on two delicately designed LoRAs~\cite{chefer2024still} we can fine-tune the model in a multi-stage scheme which will be detailed in~\cref{sec:training} on joint image-video data to synthesize high-quality video. 
We can also achieve conditional layered video generation with such a unified framework to support versatile variants.

\paragraph{Layer-wise video representation.} 
We propose a simple but effective formulation to represent different layers in videos, 
which concatenates visual embeddings of each layer including alpha-matte as a long sequence.
The following 3D attention associates between text and video and shares information among layers, contributing to inter-layer coherence.
Note that we divide foreground into RGB sequence and alpha sequence for transparency representation, thus it can be combined as ``RGBA video'' for further re-composition.

\paragraph{Layer-wise text prompts.}
To make the generated video clips separately refer to corresponding prompts,
we propose to conduct textual modification in both text input and encoded embeddings to impart layer-awareness.
Specifically, before the description of each layer encoded by $T5$~\cite{raffel2020exploring}, we attach an index number to the prompt in the format of "index number, layer description". After encoding, a learnable layer embedder projects the index number to a layer embedding of the same size as text embedding before being added to the corresponding text embedding.
All the above modifications together with position embedding link each layer to its description
by building the correspondence between pairs of textual and visual embedding explicitly and implicitly.


\paragraph{Conditional layer generation.}
\label{sec:conditional}
We can also make simple modifications to this framework to support variants of conditional layer generation, including foreground/background-conditioned generation as well as layer decomposition.
More specifically, by removing noise from the foreground clip of visual embeddings and departing it from loss calculation in training process, foreground sequence acts as condition for synthesis of remaining video segments and is only used in attention sharing for modeling relationships 
among layers.
Thus the framework becomes a foreground-conditioned video generator. The other two applications also share similar modifications and 
we demonstrate all these variations in experiments.



\subsection{Training Pipeline}
\label{sec:training}
\begin{figure*}[t]
\centering 
\includegraphics[width=0.99\linewidth]{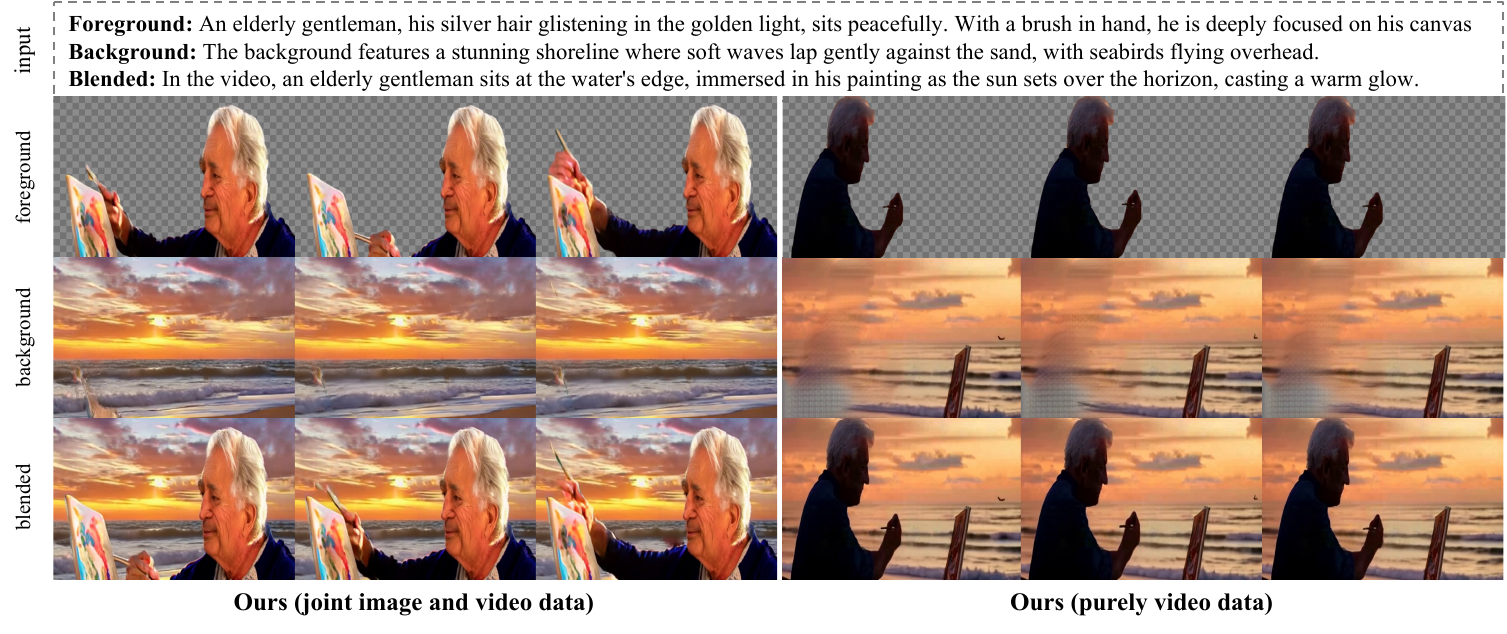} 
\caption{%
   \textbf{Ablation for training data}. We visualize the results for models trained on purely video data and joint image and video data.
   Without high-quality image data, the model tends to generate a fuzzy background with obvious blur and low fidelity, while joint image-video data training contributes to undisturbed background synthesis and a higher level of text alignment (e.g., "colorful flowers") and generation quality.
}
\label{fig:stage}
\end{figure*}


The layer-aware video generation model requires high-quality multi-layer video data, which is often difficult to obtain or construct. To address this, we propose a three-stage training scheme combining static image and dynamic video data, leveraging motion and Content LoRA to overcome data scarcity, resulting in improved inter-layer coherence, aesthetic quality, and motion dynamics.

\paragraph{First stage: base model training.}
We first train a base model on low-quality multi-layer video datasets to empower the model with initial layer-aware generation ability.
The denoising
network $\epsilon_{\theta}$ learns to predict the added noise, encouraged by an MSE loss:

\begin{align}
L(\theta):= \mathbf{E}_{t, x_{0},y,i_{l}, \epsilon} \Bigl\| \epsilon - \epsilon_{\theta} & \left(\sqrt{\bar{\alpha}_{t}} x_{0} + \sqrt{1-\bar{\alpha}_{t}} \epsilon, t, \right. \notag \\
& \left. \tau _{\theta}(y) + \tau _{l}(i_{l})\right) \Bigr\|^{2},
\label{eq:1}
\end{align}

where $x_0$ is the target video sequence and $t$ is uniformly distributed between 1 and $T$. Pre-defined 
$\bar{\alpha}_{t}$ determines the noise strength at step $t$ and $i_l$ represent the layer index, thus $\tau _{l}(i_l)$  represent the layer embedding.

To prepare the data for this stage, we first segment the foreground video sequence from raw video via SAM-Track~\cite{cheng2023segment} under the guidance of a predicted prompt list.
Then we filter out duplicate mask sequences and invalid foreground objects.
The filtered mask sequences are used to guide the video inpainting model~\cite{zhou2023propainter} in generating background videos.
At last, we utilize CogVLM2~\cite{hong2024cogvlm2} to separately caption layer-wise video data.
As a result, we acquire the $\{ foreground, alpha, background,\\ blended \}$ pairs of both textual and visual input. Note that due to the weaknesses 
of segmentation, inpainting models themselves, and poor-quality of raw videos, video frames can contain motion blur or ambiguous foreground edges, and alpha channel is binary. So there is a non-negligible quality gap between the base model output after first-stage training and current state-of-the-art video generation models as shown in~\cref{fig:stage}.


\paragraph{Second stage: Motion LoRA training.} 
A key idea to improve the quality of layered video generation is to train on high-quality but static multi-layer image data. 
To prevent loss of motion dynamics due to directly trained on static videos made by duplicated images, we propose to employ Motion LoRA as shown in~\cref{fig:pipeline} to adapt the base model to image data. Take query ($Q$) projection as an example. The internal feature z after projection becomes
\begin{align}
Q = W^Qz + Motion LoRA(z) = W^Qz + \alpha \cdot AB^Tz.
\end{align}
The implementation is conducted on all $W \in \left \{ W^Q, W^K, W^V \right \} $ 
to accommodate the motion mode between static and dynamic by adjusting the scalar $\alpha$ to 1 and 0. In other words, after optimized on duplicated frames, Motion LoRA enables the model to generate frozen videos when $\alpha$ is set to 1, accommodating the model to image data in the third stage.

For better alignment with initial feature distribution of transformer backbone trained on video dataset in the first stage, we 
sample static frame randomly from video for duplication rather than directly duplicating image data for training.
On the other hand, the quality of video data used in this stage should be considerably high since the optimized Motion LoRA
is finally employed on high-quality image data in last stage.
An alternative method to meet both requirements is to copy attainable foreground video matting data with transparency and paste on background videos to form self-made 
multi-layer video datasets. 
We need to mention that copy-pasted video datasets in this stage are only used for Motion LoRA training, thus the problem of 
incoherence due to random collage will exert no influence on the quality of layered video generation.



\paragraph{Third stage: Content LoRA training.}
With Motion LoRA adapts the model to joint image-video training,
we introduce Content LoRA on the same blocks of transformer backbone in the first two stages, which is also implemented as a multiplication of low-rank matrices:

\begin{align}
Q &= W^Qz + MotionLoRA(z) + ContentLoRA(z) \notag \\
&= W^Qz + \alpha \cdot AB^Tz + CD^Tz.
\end{align}
We optimize the Content Lora with the same diffusion reconstruction loss in \textbf{Eq.}~\ref{eq:1}, with $\alpha = 0$ for the copy-pasted videos, and $\alpha = 1$ for multi-layer dupilated images as frozen videos.
In this way, the model can learn high fidelity from image matting data without losing the motion prior of the base model.
After all three stages of training, we drop the Motion LoRA at inference time and reserve the Content LoRA for refinement, showing that this 
training pipeline helps reduce the negative effects inherited from the weak base model (e.g., defects in background filling, ambiguous foreground boundaries) and achieve transparency, high fidelity, and inter-layer harmony of multi-layer video generation.

The multi-layer image datasets used in this stage have two main sources, one is accessible multi-layer annotated datasets like MULAN~\cite{tudosiu2024mulan}, the other is image matting datasets post-processed by image inpainting and captioning. Compared to coarse video datasets used in the first stage, multi-layer image datasets possess transparent foregrounds and harmonious backgrounds without obvious motion blur or artifacts, thus
playing a decisive role in refining multi-layer generation. Note that copy-pasted videos are also used in
this stage
with a small proportion of joint data, which has a weak influence on generation coherence but effectively improves fidelity and prevents the model from loss of dynamic property.
\begin{figure}[h]
\centering 
\includegraphics[width=1\linewidth]{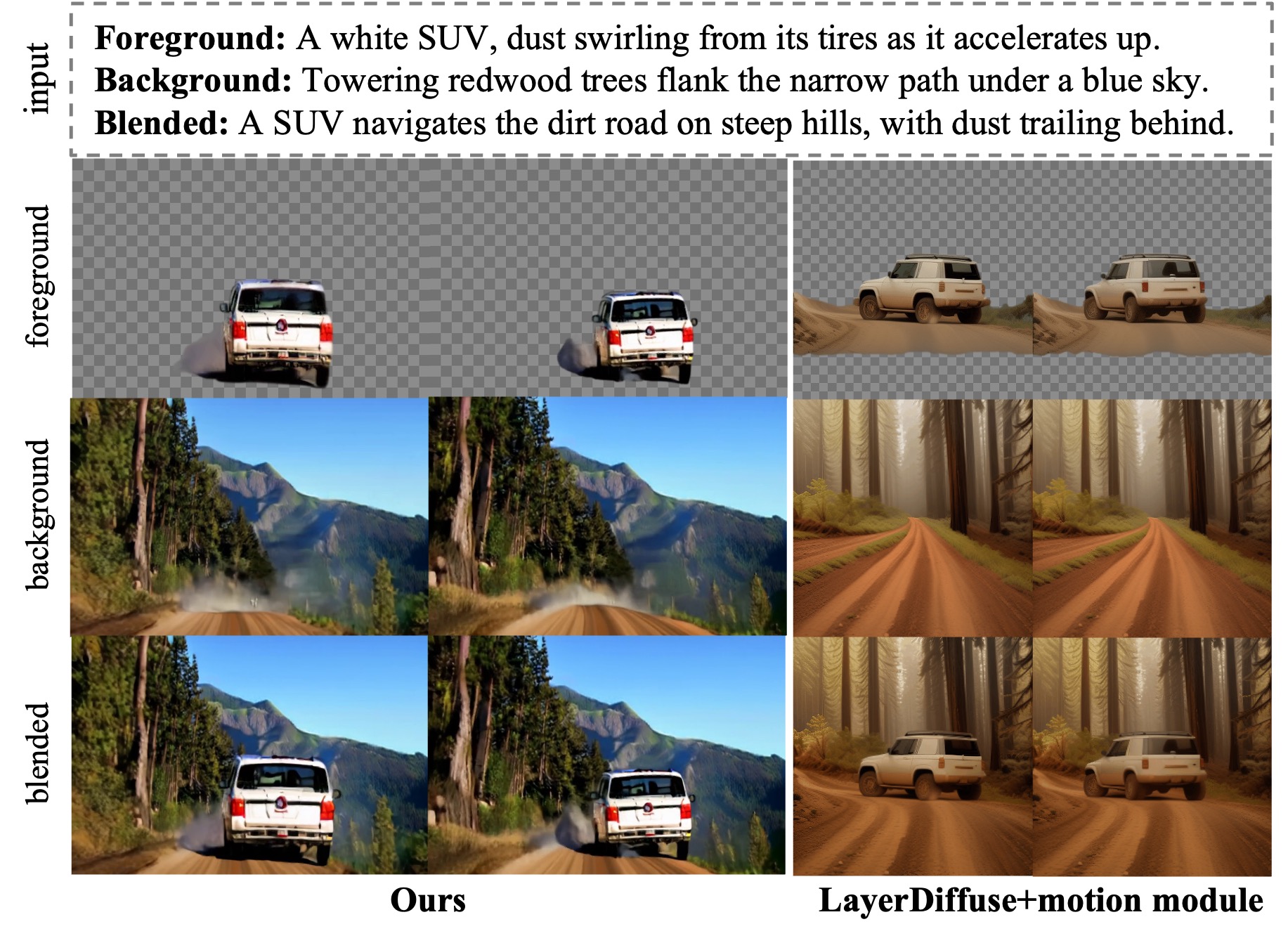} 
\caption{%
   \textbf{Qualitative comparison for multi-layer video generation} with generation then animation pipeline, i.e., composition of LayerDiffuse~\cite{zhang2024transparent} and motion module~\cite{guo2023animatediff}, where \method 
   achieves better layer-level coherence and clearer separation of layers.
}
\label{fig:comp}
\end{figure}
\section{Experiments}
\label{sec:exp}


\begin{figure*}[t]
\centering 
\includegraphics[width=1\linewidth]{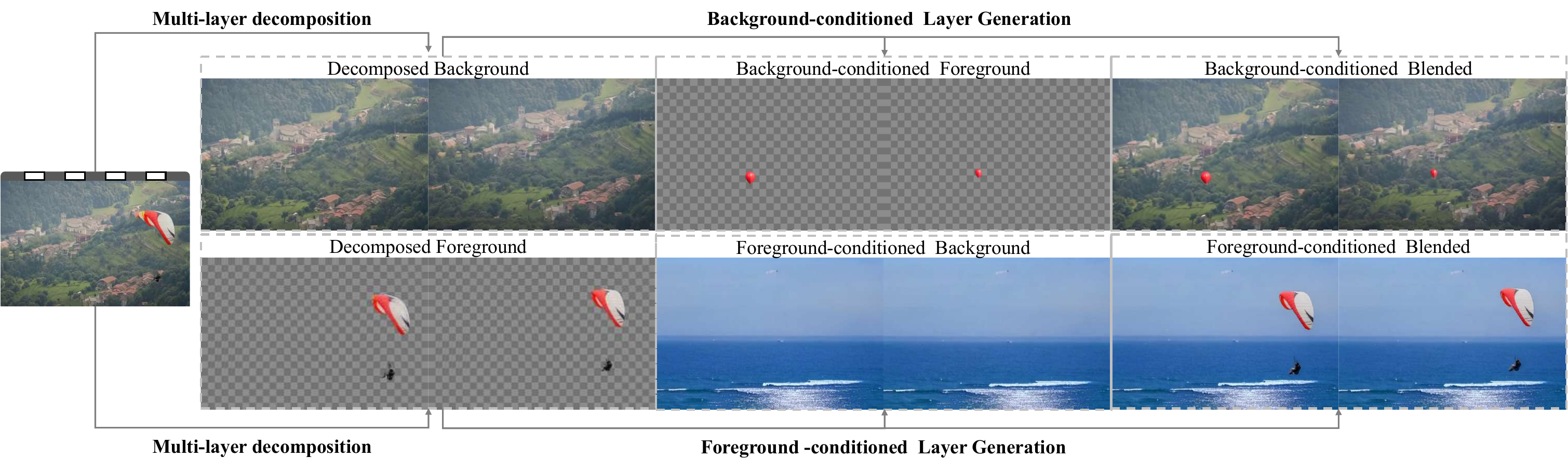} 
\caption{%
   \textbf{Qualitative results for iterative multi-layer video generation}, which iteratively implement multi-layer decomposition and conditioned layer generation to recompose video assets.
}
\label{fig:iterative}
\end{figure*}

\subsection{Implementation Details}
\paragraph{Training configurations.}
We implement \method based
on the T2V model CogVideoX~\cite{yang2024cogvideox} with 2B parameters.
We sample training videos of each layer with 16 frames, thus the frame number becomes 64 after concatenation,
and resize each frame to 480 × 720 pixels as CogVideoX~\cite{yang2024cogvideox} does. 
We only use a simple MSE loss to train the model in all three stages and adopt Adam optimizer with a learning rate of $1e^{-4}$
for base model fine-tuning, $1e^{-3}$ for Motion LoRA and $5e^{-3}$ for Content LoRA training.
Motion LoRA and Content LoRA are also attached on the trainable 1/6 transformer blocks in base model.
The model is optimized on 8 NVIDIA A800 GPUs with batch size of 12 for each GPU in training, while in the inference process, 
number of sampling steps is set as 50 with classifier-free guidance scale of 6.

\paragraph{Coarse multi-layer video dataset construction.}
First, the prompt list used for foreground segmentation is produced by Recognize Anything~\cite{zhang2024recognize}, composed of possible foreground subjects of the video. With the generated foreground masks, the filtering is then conducted according to the similarity among all masks to avoid duplicated foreground objects. Qwen-VL~\cite{bai2023qwen} further helps to check whether the segmented sequence is a ``real'' foreground to exclude samples commonly viewed as background like sky, lake, etc. Based on the above steps, we completed the construction of coarse layered video dataset.

\paragraph{Evaluation protocols.}
We conduct user studies and utilize four metrics from
VBench~\cite{huang2024vbench} for quantitative assessment:
Frame Consistency measures consistency between adjacent frames;
Aesthetic Quality considers artistic value;
Text Alignment reflects both semantics and
style consistency on text prompts; 
Dynamic Degree depends on whether a video contains large motion or not.
The evaluation prompt sets are specially designed with help of GPT-4 and CogVLM2~\cite{hong2024cogvlm2} as text inputs. Both qualitative and quantitative comparisons are carried out on those prompt sets. 
Since we have different prompts for foreground, background, and blended videos, 
the performance is assessed separately on each category.
The subtask of text-guided layered video decomposition which encompasses both segmentation and inpainting, enables the reconstruction of foreground and background regions occluded by one another. This distinguishes it from conventional segmentation tasks and makes traditional segmentation metrics unsuitable for its evaluation.
As a result, the same metrics except for Dynamic Degree as those used for generation are applied to the decomposed foreground and background videos for quantitative evaluation. The raw videos for decomposition are from DAVIS~\cite{perazzi2016benchmark} comprised of 50 sequences.
\subsection{Comparisons with Existing Alternatives}
\label{sec:comp} 
It is worth clarifying that since we are studying a novel problem, 
there is no prior work operating under the exact same
setting to the best of our knowledge. We hence compare with  a multi-stage alternative achieving the same goal, which
also demonstrates that the formulation and overall
pipeline is one of the core contributions of our work.
We compare
\method with the alternative via 
both qualitative analysis and user study on the core functionality of multi-layer video generation.

\begin{table*}[t]
  \setlength{\belowcaptionskip}{0cm}
  \footnotesize
  \renewcommand{\arraystretch}{0.85} 
  \caption{\textbf{Quantitative analysis for model framework and training data of multi-layer generation}. Two groups of comparison are included, one is between our model trained without (top row) or with (bottom row) image data, and the other is among different architectures including our framework, LayerDiffuse~\cite{zhang2024transparent}+motion module~\cite{guo2023animatediff}, and ``Channel-concatenate'' architecture (first three rows). Here, “FG”, "BG", and "BL" refer to foreground, background, and blended layer.
}
  
  \label{tab:gen}
  \centering
  \scriptsize
  \resizebox{0.99\textwidth}{!}{ 
    \scriptsize 
    \begin{tabular}{lccc|ccc|ccc|ccc}
      \toprule
      \multirow{2}{*}{Methods} & \multicolumn{3}{c}{Dynamic Degree} & \multicolumn{3}{c}{Aesthetic Quality} & \multicolumn{3}{c}{Text Alignment}
      & \multicolumn{3}{c}{Frame Consistency} \\
      \cmidrule(lr){2-4} \cmidrule(lr){5-7} \cmidrule(lr){8-10} 
      \cmidrule(lr){11-13} 
      & FG & BG & BL & FG & BG & BL & FG & BG & BL & FG & BG & BL \\
      \midrule
      Ours (purely video data) & \underline{0.32} & - & \underline{0.60} & 0.4189 & \underline{0.4364} & \underline{0.4959} & 0.1757 & \underline{0.1861} & \textbf{0.2428} & 0.9612 & 0.9643 & 0.9608 \\
      LayerDiffuse+motion module & 0.05 & - & 0.36 & \underline{0.4487} & 0.3701 & 0.4868 & \underline{0.1841} & 0.1695 & 0.1674 & \textbf{0.9669} & \textbf{0.9888} & \textbf{0.9747} \\
      Channel-concatenate & 0.02 & - & 0.01 & 0.3454 & 0.2899 & 0.3871 & 0.1348 & 0.0404 & 0.1585 & 0.9493 & 0.9543 & 0.9549 \\
      \textbf{Ours (joint image-video data)} & \textbf{0.62} & - & \textbf{0.66} & \textbf{0.4859} & \textbf{0.5753} & \textbf{0.5742} & \textbf{0.1972} & \textbf{0.2312} & \underline{0.2350}& \underline{0.9621} & \underline{0.9777} & \underline{0.9633} \\
      \bottomrule
    \end{tabular}
  }
\end{table*}

\begin{table}[t]
\caption{%
    \textbf{User study for multi-layer video generation on \method and existing alternatives}. “Quality” and “T-A” measure overall synthesis quality and text alignment respectively, “FG”, “BG”, and “BL” refer to the detailed evaluation of the foreground, background, and blended video.
}
\label{tab:userstudy}
\centering
\footnotesize
\setlength{\tabcolsep}{1.5pt}
\resizebox{0.48\textwidth}{!}{ 
\begin{tabular}{lccccc}
\toprule
{Methods} & {Aesthetic~($\uparrow$)} & {FG~($\uparrow$)} & {BG~($\uparrow$)} & {BL~($\uparrow$)} & {T-A~($\uparrow$)} \\
\midrule
{Channel-concatenate} & {40.69} & {34.85} & {42.13} & {38.23} & {40.61} \\
{LayerDiffuse+motion module} & {46.58} & {56.07} & {56.60} & {45.92} & {37.92} \\
{Ours (purely video data)} & {\underline{73.58}} & {\underline{67.33}} & {\underline{56.81}} & {\underline{74.68}} & {\underline{75.57}} \\
{\textbf{Ours (joint image-video data)}} & {\textbf{89.15}} & {\textbf{91.75}} & {\textbf{94.46}} & {\textbf{91.17}} & {\textbf{95.90}} \\
\bottomrule
\end{tabular}
}
\end{table}

\begin{table}[t]
  \setlength{\belowcaptionskip}{0cm}
  \renewcommand{\arraystretch}{0.9} 
  \caption{
  \textbf{Quantitative analysis for model framework and training data of multi-layer video decomposition}. Two groups of comparison are included, one is between our model trained without (top row) and with (bottom row) images, and the other is between different architectures of our base model (top row), and ``Channel-concatenate'' (middle row).
}
  
  \label{tab:seg}
  \centering
  \resizebox{0.49\textwidth}{!}{ 
    \begin{tabular}{lcc|cc|cc}
      \toprule
      \multirow{2}{*}{Methods} & \multicolumn{2}{c}{Aesthetic Quality} & \multicolumn{2}{c}{Text Alignment} & \multicolumn{2}{c}{Frame Consistency} \\
      \cmidrule(lr){2-3} \cmidrule(lr){4-5} \cmidrule(lr){6-7} 
      & FG & BG & FG & BG & FG & BG \\
      \midrule
      Ours (purely video data) & \underline{0.3973} & \underline{0.2822} & \underline{0.1604} & \underline{0.0436} & \underline{0.9371} & \underline{0.9370} \\
      Channel-concatenate & 0.3265 & 0.2782 & 0.1400 & 0.0388 & 0.9280 & 0.9301 \\
      \textbf{Ours (joint image and video data)} & \textbf{0.4240} & \textbf{0.3020} & \textbf{0.1872} & \textbf{0.0471} & \textbf{0.3973} & \textbf{0.9424} \\
      \bottomrule
    \end{tabular}
  }
\end{table}

\paragraph{Qualitative analysis.}
Specifically, we compare our method with a generation then animation baseline.
The alternative architecture starts from AnimateDiff~\cite{guo2023animatediff}, a prominent T2V design where the
video model is built over a text-to-image (T2I) model inflation. We replace the base T2I model 
in AnimateDiff~\cite{guo2023animatediff} with LayerDiffuse~\cite{zhang2024transparent} and try to synthesize multi-layer videos by inflating it with motion module~\cite{guo2023animatediff}.
However naive plugging of motion module
doesn't work for LayerDiffuse~\cite{zhang2024transparent} since 
a conflict exists between the attention sharing in dimension of layer and time. Thus we
modify the architecture, where we separately conduct temporal attention within each layer of video frames and layered attention 
among three layer sequences to resolve this conflict.
Although multi-layer video generation is
achieved, the composition of two attention mechanisms without tuning leads to lack of inter-layer coherence as shown in~\cref{fig:comp}. Besides, our model also demonstrates significant advantages in motion dynamics and text alignment in comparison.
Moreover, our model completes the entire pipeline in a single pass, allowing layer-wise
interaction for generation coherence and showcasing 
text-aligned distinguishment between layers.

\paragraph{User study.}
Although qualitative comparison has demonstrated
significant advantages of multi-layer video generation over ad-hoc
solution like combination of pipelines, it has limitations in thoroughly evaluating the model.
In the absence of an appropriate metric
to evaluate inter-layer consistency and coherence, we resort to a user
study for further quantitative comparison.
{30} annotators are required to rate the generated videos
from five key aspects. Artistic quality considers the overal quality of three layered videos like the richness and harmony of colors and layout. Foreground quality evaluates the completeness and clarity of the foreground layer;
Background quality measures completeness and adherence to the objective of the physical world of the background video.
Blended quality assesses harmony and naturalness of blended video while text alignment checks whether the motion and content of videos follow each text
description. The results are shown in \cref{tab:userstudy}. We select 25 groups of descriptions including subjects like humans, animals and still objects from prompt sets, then the participants are invited to rank the 25 groups of generated results by \method and alternative pipelines (generated results of our model trained by purely video data and another framework is also included for ablative analysis, which will be further explained in ~\cref{sec:abla}), then the preference scores for each group rated from 1~(worst) to 4~(best) are summed up with a full mark of 100 as the testing metric.
The second row and bottom row of ~\cref{tab:userstudy} show that  \method performs significantly better text consistency and overall quality over the alternative pipeline.
\subsection{Ablation Studies}
\label{sec:abla}
We ablate crucial designs of our method from two broad perspectives, the training mechanism and the framework architecture.

\paragraph{Training mechanism.}
As demonstrated in~\cref{fig:stage}, before applying Motion LoRA and Content LoRA for further tuning, the generation results suffer from unclear boundaries
of foreground and severe blurring of background.
While after joint image-video training based on two LoRAs, 
the beauty value of foreground is improved and the background is out of fuzziness.
The quantitative comparison results on two subtasks, i.e., multi-layer generation and decomposition, are separately listed in \cref{tab:gen} and \cref{tab:seg} (top row and bottom row), both consistently confirming the effectiveness of our training mechanism.
As for generation, 
scores of Frame Consistency 
reflect that the introduction of joint data allows high-level of appearance consistency of adjacent frames,
Image data also contributes to
artistic and beauty value perceived by humans according to Aesthetic Quality.
Reported text alignment scores witness increasing consistency of semantics on prompts in foreground and background, but decent performance of text alignment in blended videos.
It is explainable that joint data, which has a clearer and more distinct separation between foreground and background, will help improve the association between text and video of foreground and background. However, the textual alignment in blended videos may be sacrificed due to differences in distribution in text-video pairs and text-image pairs.
Dynamic Degree is introduced 
as an additional reference for frame consistency since static video will also score well in this temporal metric.
In other words, with the dynamic degrees increase in the foreground and blended video, the frame consistency produced by our joint data training model is still proved to be improved, showcasing that the contents do remain consistent throughout the whole video. Note that the prompts of background often describe a nearly static scene, thus the dynamic degree is not measured on background videos.

User study also demonstrates obvious superiorities of our training strategies as shown in the top row and bottom row of~\cref{tab:userstudy}, 
We also carry out a similar ablation analysis on the multi-layer decomposition task as in top and bottom row of~\cref{tab:seg}
Frame consistency and Text Alignment are all improved, verifying the benefits
of joint data training in making layer separation clearer and semantically aligned.
We also find that the quality of aesthetics undergoes considerable improvement.

\paragraph{Overal framework.}
We analyze the generated results of different model architectures with the conclusion that our framework design outperforms all other explorations.
The first alternative architecture is training-free by composition of LayerDiffuse~\cite{zhang2024transparent} and motion module~\cite{guo2023animatediff} as detailed in \cref{sec:comp}. It only works for video generation, unable to accomplish the decomposition task.
The second alternative architecture is more similar to ours where the video
sequences are not concatenated in dimension of time but in dimension of layer.
It takes 6D video tensors $z \in \mathbb{R}^{b \times l \times t \times c \times h \times w}$
as input, where b stands for batch axis, $l$ and $t$ represent layer axis and time axis respectively.
When the internal feature maps go through transformer blocks, the layer axis is reshaped into the channel axis $c$ to get expanded feature maps $z \in \mathbb{R}^{b \times t \times (l \times c )\times h \times w}$.
With the following layer projecting the tensors to original size, such a framework
is also ready to complete the generation and decomposition tasks after fine-tuning, which is denoted as ``Channel-concatenate''.

We compare \method trained by purely video data with the training-free pipeline and the ``Channel-concatenate" architecture trained on same datasets for a fair comparison. ``LayerDiffuse + motion module'' performs worse than our model on all metrics except for frame consistency due to particularly low dynamic degree. ``Channel-concatenate'' architecture fails to show promising performance in both setting, possibly due to the difficulty for lightweight projection layer in preserving sufficient visual information among layers as showcased in~\cref{tab:gen} and~\cref{tab:seg}.
In contrast, even \method purely trained on video data (top row) owns a dominant superiority over other alternative architectures for both generation and decomposition, joint data training mechanism further improves the generation quality across all aspects, demonstrating the efficacy of our whole model design.
\subsection{Versatile Variants}
\label{sec:variants}
\begin{figure*}
\centering 
\includegraphics[width=0.95\linewidth]{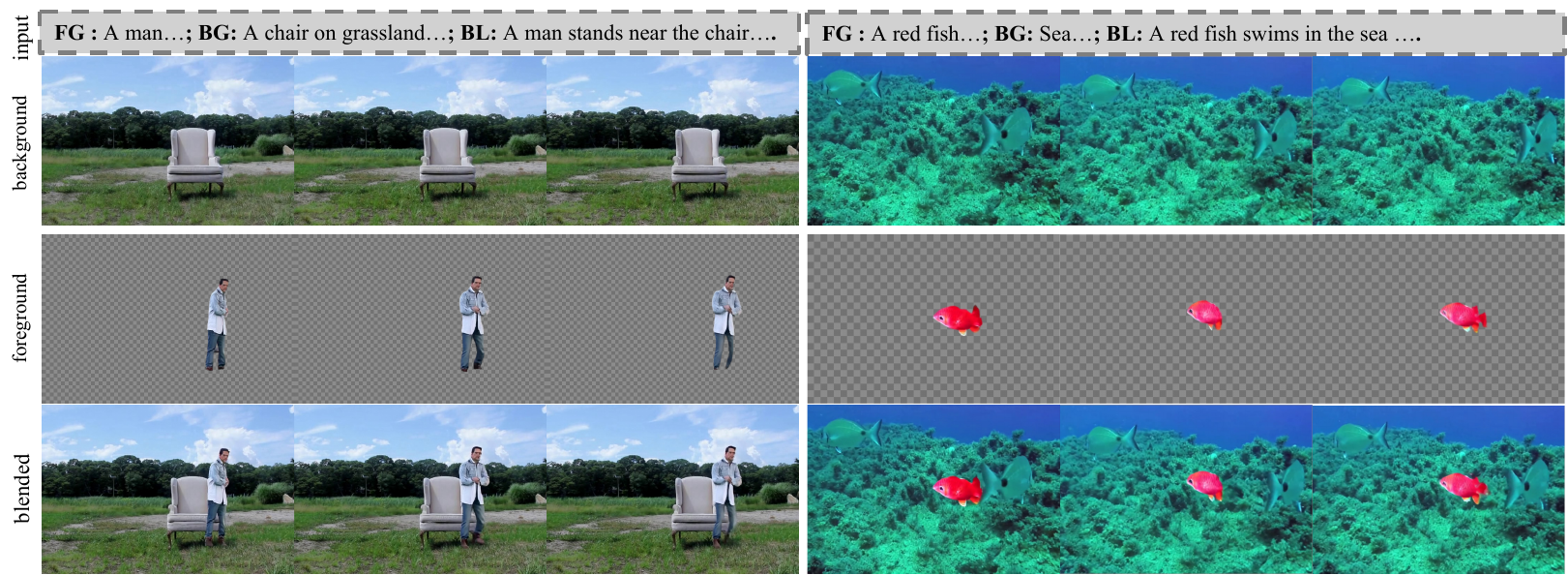} 
\caption{%
   \textbf{Demonstrations for background-conditioned layer generation}, where we take three layer-wise descriptions and a background sequence as input (top row) and show generated results of foreground (middle row) and blended video (bottom row).
}
\label{fig:bg2fg}
\end{figure*}

\begin{figure*}
\centering 
\includegraphics[width=0.95\linewidth]{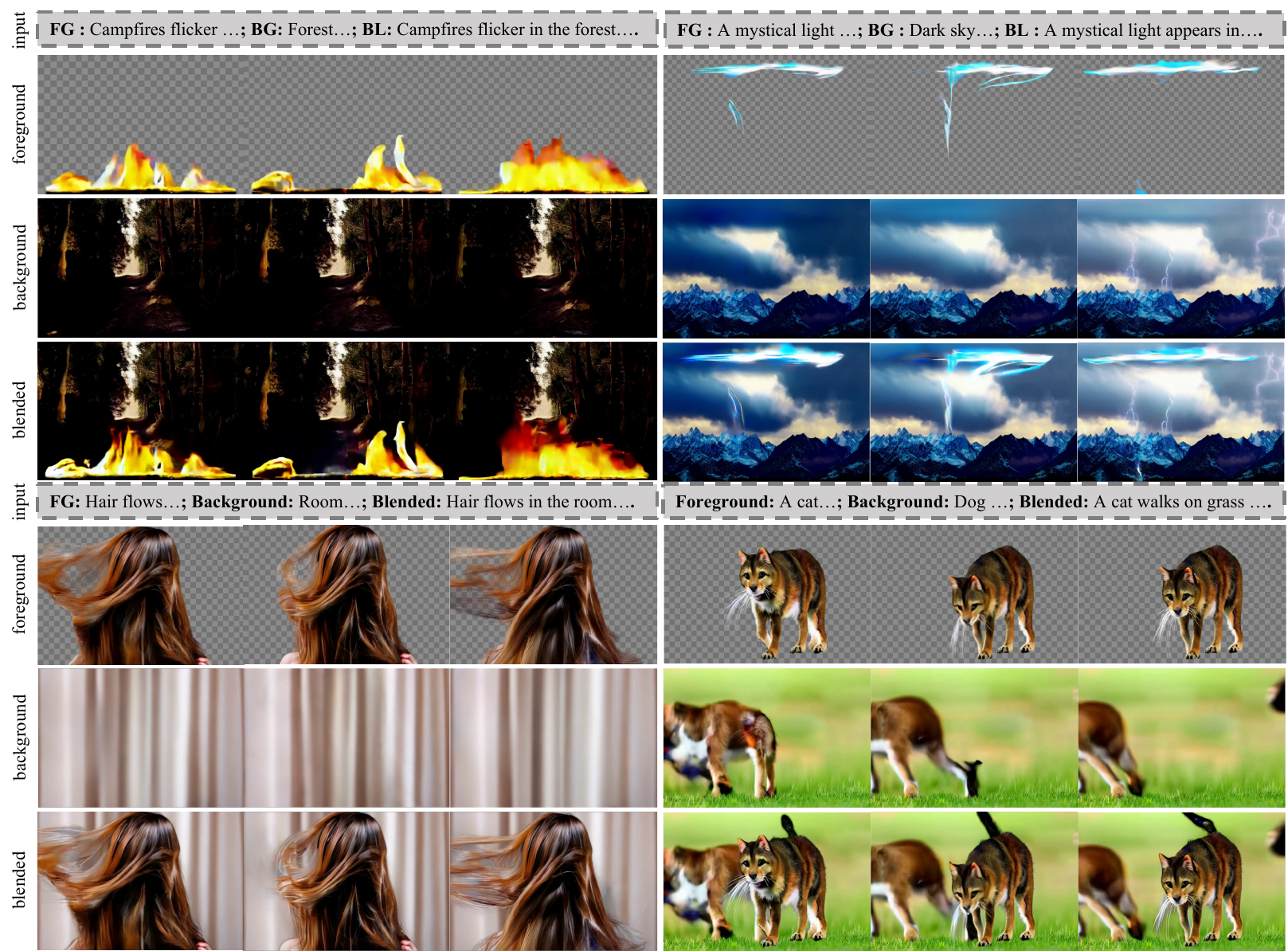} 

\caption{%
   \textbf{Demonstrations for multi-layer video generation.}
   For each example, we take three layer-wise descriptions as input and show generated results of foreground (top row), background (middle row)
   and blended video (bottom row). 
}
\label{fig:gen}
\end{figure*}

\begin{figure*}[t]
\centering 
\includegraphics[width=0.95\linewidth]{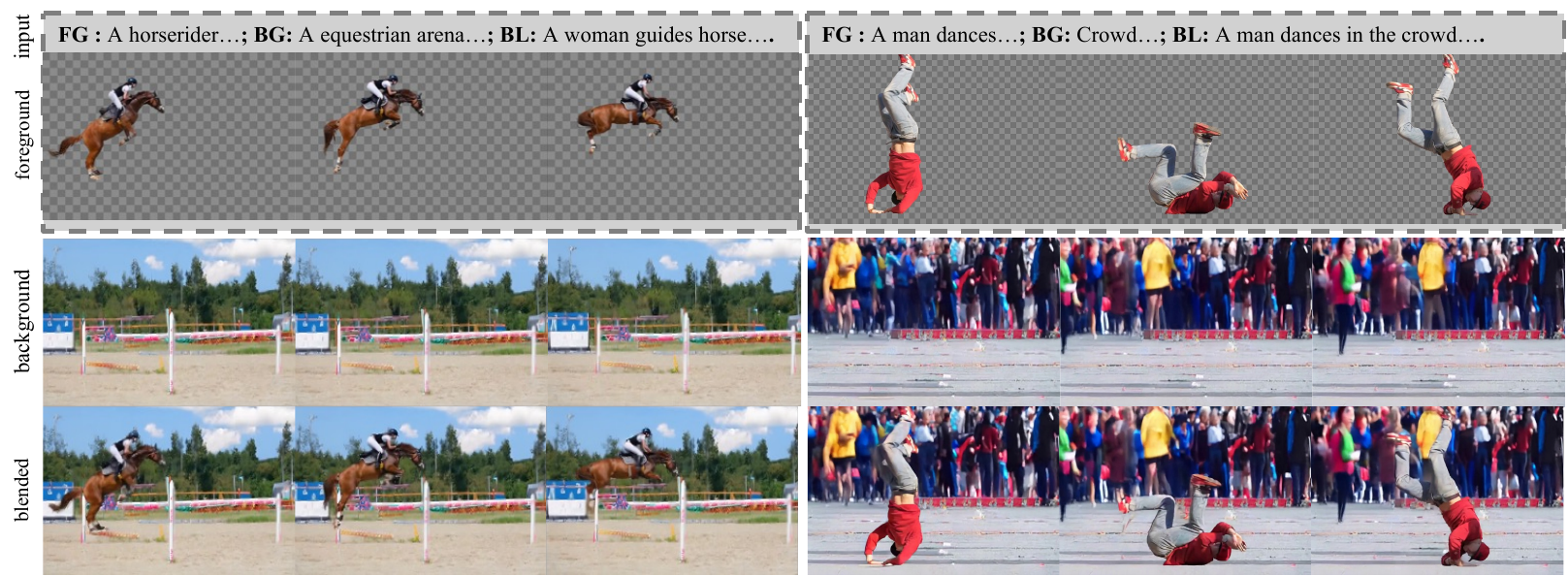} 

\caption{%
   \textbf{
   Demonstrations for foreground-conditioned layer generation}, where we take three layer-wise descriptions and a foreground sequence as input (top row) and show generated results of background (middle row) and blended video (bottom row).
}
\label{fig:fg2bg}
\end{figure*}

\begin{figure*}
\centering 
\includegraphics[width=0.95\linewidth]{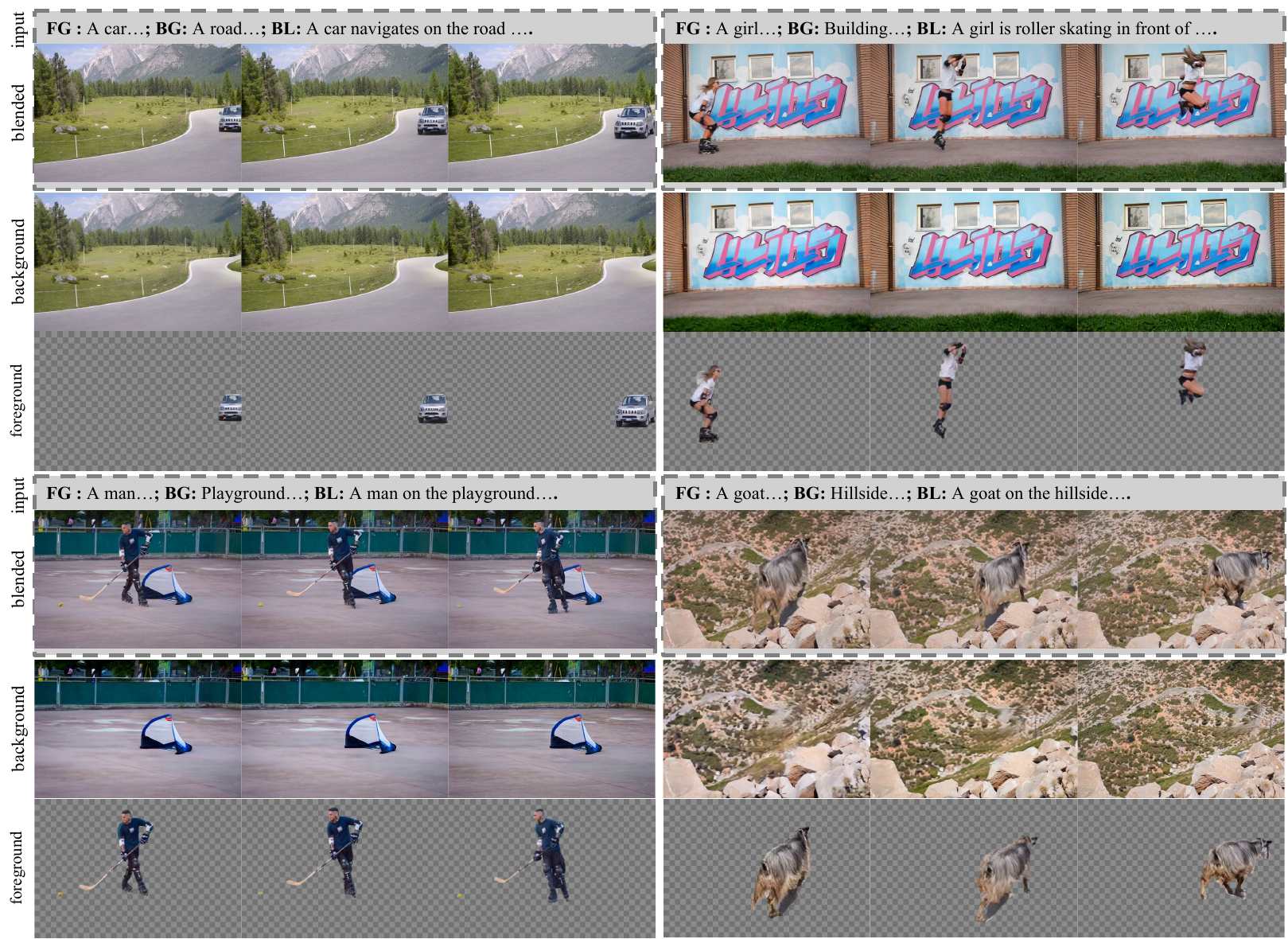} 

\caption{%
   \textbf{
   Demonstrations for multi-layer video decomposition}, where we take three layer-wise descriptions and a blended sequence as input (top row) and show generated results of background (middle row) and foreground video (bottom row).
}
\label{fig:seg}
\end{figure*}

A set of generated multi-layer videos are shown in~\cref{fig:gen}.
Using three descriptions corresponding to each layer, 
\method shows generation capability in both text alignment and artistic quality. 
For versatile variants, we prove our model's ability by presenting samples in~\cref{fig:fg2bg,fig:bg2fg,fig:seg}, demonstrating 
\method's extended capabilities for harmonious foreground or background-conditioned layer generation and multi-layer video decomposition.
We also demonstrate results of iterative layered video generation in~\cref{fig:iterative}, by conduct multi-layer decomposition followed by conditioned multi-layer video generation with a raw video as input, showing our model's ability in video recomposition, and we have reasons to believe its potential for future applications.

\section{Conclusion}

We introduce \method, a unified multi-layer video generation model guided by textual input and optional video conditions. \method is built upon a novel framework with inherent layer awareness, enabling effective multi-layer generation, decomposition, and foreground/background-conditioned video synthesis. 
To address the challenges of limited training data, we propose a joint training strategy that integrates both video and image data. This is achieved using two specially designed components: Motion LoRA, which adjusts dynamic motion synthesis, and Content LoRA, which enhances layer separation and improves overall visual quality. 
Our pipeline demonstrates impressive performance, making it well-suited for various applications in video creation and editing.

\paragraph{Limitations.} Our limitation lies in the model’s inability to support multi-layer generation with a variable number of layers. We expect to address this in future work by enabling video generation with a flexible number of layers, allowing for more dynamic and complex scene compositions.

\begin{acks}
This work is supported by the National Natural Science Foundation of China (No. 62441615, 62422606, 624B2124) and Damo Academy through Damo Academy Research Intern Program. 
\end{acks}
%
%
%
%

\bibliographystyle{ACM-Reference-Format}
\bibliography{sample-bibliography}

\end{document}